\documentclass[10pt,twocolumn,letterpaper]{article}

\usepackage{cvpr}              
\usepackage{algorithm}
\usepackage{algpseudocode}
\usepackage{float}

\usepackage{multicol}
\usepackage{multirow}
\usepackage{makecell}

\definecolor{cvprblue}{rgb}{0.21,0.49,0.74}
\usepackage[pagebackref,breaklinks,colorlinks,allcolors=cvprblue]{hyperref}

\usepackage{multirow} 
\def\paperID{4620} 
\def\confName{CVPR}
\def\confYear{2025}

\title{Attention Distillation: A Unified Approach to Visual Characteristics Transfer}

\author{Yang Zhou \qquad Xu Gao \qquad Zichong Chen  \qquad Hui Huang\thanks{Corresponding author.}\\
	Visual Computing Research Center, CSSE, Shenzhen University\\
}

\begin{document}
 \include{_paper}

 \newpage
 \appendix
 \clearpage


\begin{algorithm*}[!ht]
    \caption{Content-preserving Optimization (For Style and Appearance Transfer)}
    \label{algo:optimization}
    \begin{algorithmic}[1] 
    \State \textbf{Input:} Style image $I^s$, content image $I^c$, learning rate $\eta$, content loss weight $\lambda$.
    \State \textbf{Output:} Optimized image $I$.
        \State $z^s, z^c \gets \mathcal{E}(I^s), \mathcal{E}(I^c)$  \Comment{Convert the input images to latent space}
        \State Initialize $z \gets z^c$  \Comment{Start with the content latents}
        \For{$t$ = $T$, $T-1$, ..., $1$}
            \State $\{Q_c, K_c, V_c\} \gets \epsilon_\theta(z^c, t, \emptyset)$\Comment{Extract self-attention features from the UNet}
            \State $\{Q_s, K_s, V_s\} \gets \epsilon_\theta(z^s, t, \emptyset)$
            \State $\{Q, K, V\} \gets \epsilon_\theta(z, t, \emptyset)$
            \State $\mathcal{L}_{\mathrm{content}}=\|Q-Q_c\|_1$  \Comment{Calculate the content loss}
            \State $\mathcal{L}_{\mathrm{AD}}=\|\mathrm{Self}\text{-}\mathrm{Attn}(Q,K,V)-\mathrm{Self}\text{-}\mathrm{Attn}(Q,K_{s},V_{s})\|_{1}$  \Comment{Calculate the style loss}
            \State $\mathcal{L}_{\mathrm{total}}=\mathcal{L}_{\mathrm{AD}}+\lambda\mathcal{L}_{\mathrm{content}}$  \Comment{Total loss}
            \State $z \gets z - \eta \nabla_{z} \mathcal{L}_{\mathrm{total}}$  \Comment{Gradient descent step}
        \EndFor
    \State $I\gets \mathcal{D}(z)$  \Comment{Decode the latents to image space}
    \State \textbf{Return:} $I$.
    \end{algorithmic}
\end{algorithm*}

\begin{algorithm*}[!ht]
    \caption{Attention Distillation Guided Sampling (For Style-specific Text-to-Image Generation)}
    \label{algo:sampling}
    \begin{algorithmic}[1] 
    \State \textbf{Input:} Style image $I^s$, text prompt $y$, learning rate $\eta$, optimization steps $M$.
    \State \textbf{Output:} Generated image $I$.
        \State $z^s \gets \mathcal{E}(I^s)$  \Comment{Convert the input images to latent space}
        \State Initialize $z_T \sim \mathcal{N}(0, 1)$ \Comment{Start with random noise}
        \For{$t$ = $T$, $t-1$, ..., $1$}
            \State $z_{t-1} \gets \mathrm{Sampling}(z_t, t, \epsilon_\theta(z_t, t, y))$ \Comment{Diffusion Sampling}
            \State $z^s_{t-1} \gets \sqrt{\bar{\alpha}_{t-1}}z^s+\sqrt{1-\bar{\alpha}_{t-1}}\epsilon$, $\epsilon\sim\mathcal{N}(0,1)$ \Comment{Add noise to the style image latents}
            \State $\{Q, K_s, V_s\} \gets \epsilon_\theta(z^s_{t-1}, t-1, \emptyset)$ \Comment{Extract self-attention features from the UNet}
            \State $z_{t-1} = \mathrm{AdaIN}(z_{t-1}, z^s_{t-1})$ \Comment{Modulate the variance and mean}
            
            \For{$m$ = $1$, ..., $M$}
                \State $\{Q, K, V\} \gets \epsilon_\theta(z_{t-1}, t-1, \emptyset)$
                \State$\mathcal{L}_{\mathrm{AD}}=\|\mathrm{Self}\text{-}\mathrm{Attn}(Q,K,V)-\mathrm{Self}\text{-}\mathrm{Attn}(Q,K_{s},V_{s})\|_{1}$\Comment{Calculate the style loss}
                \State$z_{t-1} \gets z_{t-1} - \eta \nabla_{{z_{t-1}}} \mathcal{L}_{\mathrm{AD}}$ \Comment{Gradient descent step}
            \EndFor
        \EndFor
    \State $I\gets \mathcal{D}(z_0)$  \Comment{Decode the latents to image space}
    \State \textbf{Return:} $I$
    \end{algorithmic}
    
\end{algorithm*}

\begin{figure*}[!hbt]
    \centering
    \includegraphics[width=\linewidth]{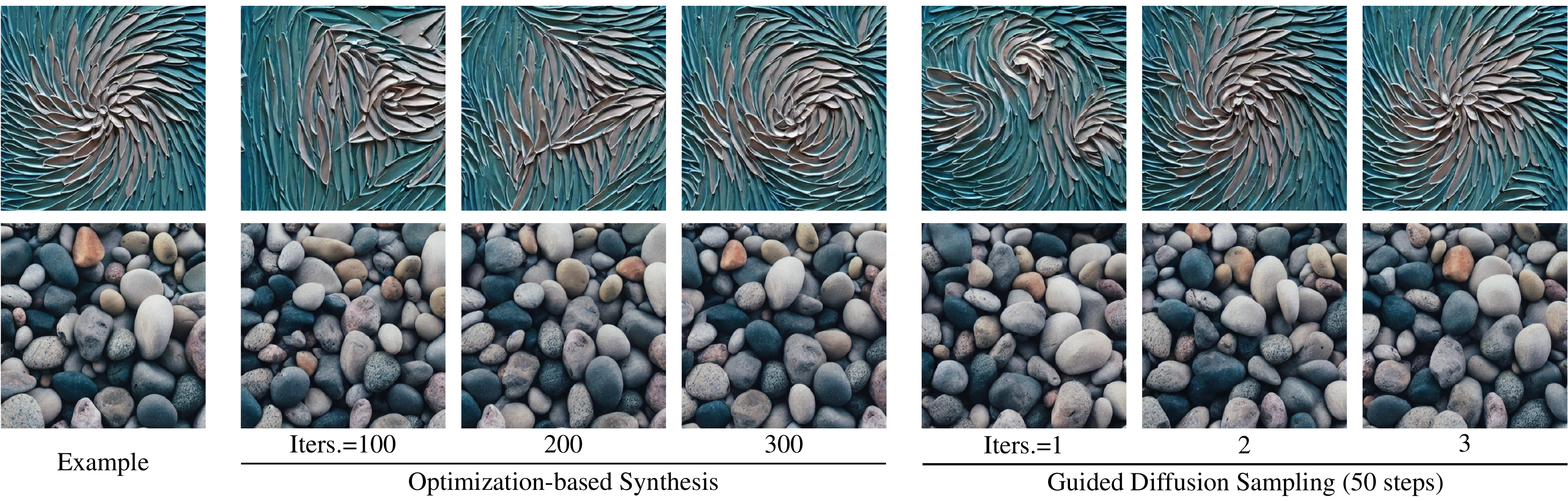} 
    \caption{Comparison between optimization-based and sampling-based approaches with attention distillation.}
    \label{fig:cmp_two}
\end{figure*}

\section{Algorithm}
We build our method on the pretrained Stable Diffusion models. Algorithm~\ref{algo:optimization}, using style transfer as an example, outlines our content-preserving optimization approach with attention distillation loss. For attention distillation guided sampling, we take style-specific text-to-image generation as an example and describe our approach in Algorithm~\ref{algo:sampling}. We denote the encoder and decoder of VAE as $\mathcal{E}(\cdot)$ and $\mathcal{D}(\cdot)$, respectively, and use $\epsilon_\theta(\cdot)$ to represent the denoising network. In Algorithm~\ref{algo:sampling}, $\mathrm{Sampling}(\cdot)$  refers to a diffusion sampling step from $z_t$ to $z_{t-1}$, and $\mathrm{AdaIN}(\cdot, \cdot)$~\cite{AdaIn} refers to modulate the variance and mean of the features to boost stylization.

\section{Implementation Details}

We implemented our approach using the PyTorch framework, applying mixed precision to save time and memory costs. For style-specific text-to-image generation, we use SDXL~\cite{Podell2023SDXLIL}; for other tasks, we employ Stable Diffusion v1.5~\cite{Rombach2021HighResolutionIS}. Following recent works~\cite{cao2023masactrl, zhou2024generating, jeong2024visual}, we extract attention features from the last six self-attention layers of U-Net to compute attention distillation loss. For comparison, we use the publicly available implementations of all baseline methods and adhere to their suggested configurations. All experiments are conducted on a single NVIDIA RTX 6000 Ada GPU. We use a fixed learning rate (0.05) for the Adam optimizer, except for style-specific text-to-image generation (0.015). In the following, we specify the detailed configurations for each task.

\paragraph{Style/Appearance Transfer.} We initialize the target latent using the content/structure image. The content loss is computed with the $Q$ features from the last 6 self-attention layers. The content loss weight, $\lambda$, is set to 0.25 for style transfer and 0.2 for appearance transfer, respectively. By default, We optimize the target latent over 200 iterations. All experiments are conducted to generate images at a resolution of 512x512. The time to synthesize an image takes about 30 seconds, with our optimization in latent space.

\paragraph{Style-Specific Text-to-Image Generation.} We generate images at a resolution of 1024x1024 using SDXL. The sampling is conducted over 50 steps using DDIM sampling, with a scale set to 7 for classifier-free guidance. At each sampling step, We perform 2 iterations of latent optimization utilizing attention distillation loss. The whole process takes no more than 30 seconds. The learning rate of the Adam optimizer is set to 0.015 by default.



\begin{figure}
    \centering
    \includegraphics[width=\linewidth]{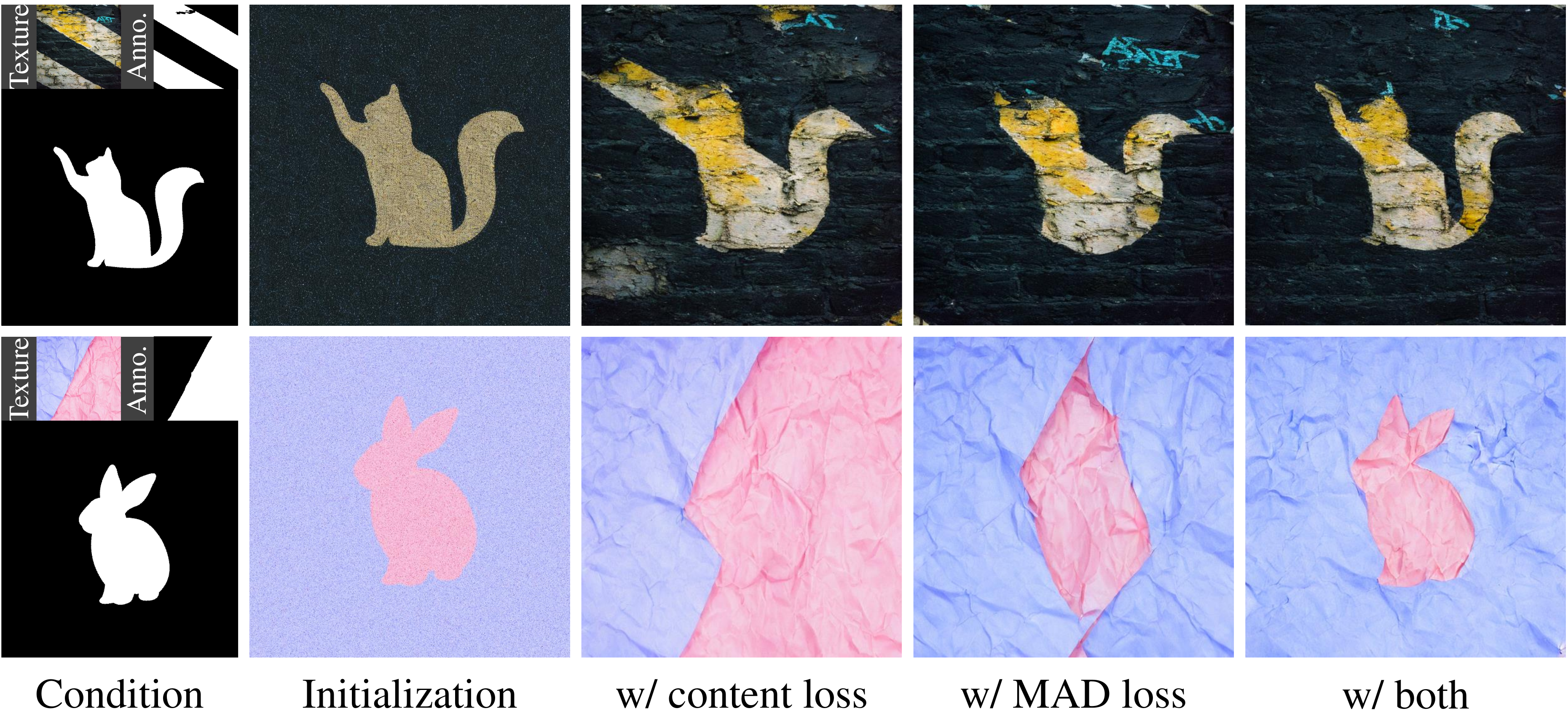} 
    \caption{Ablation of losses used for controlled texture synthesis.}
    \label{fig:abl_ctrltex}
\end{figure}

\paragraph{Controlled Texture Synthesis.} For the mask-controlled texture synthesis, images are resized to resolution 512$\times$512, and synthesized in the optimization manner. The optimization performs 200 iterations by default.  We adopted the same initialization strategy as GCD~\cite{zhou2023neural}, where we fill the target segmentation map with random pixels drawn from the semantically corresponding region of the source texture. However, the low spatial resolution of features from U-Net makes the Masked AD loss inadequate for precise spatial control, as shown in Fig.~\ref{fig:abl_ctrltex}. To address this, we utilize the $Q$ features from the initialization image to compute the content loss with a content weight $\lambda$ of 0.15. Introducing content loss leads to precise spatial alignment without compromising texture quality. For the layout control task as Self-Rectification~\cite{zhou2024generating}, we directly use the color layout as the content image to compute content loss.

\paragraph{Texture Expansion.} In this task, the example textures are resized to 512$\times$512. The results are generated using attention distillation guided sampling for efficiency. We use MultiDiffusion~\cite{bar2023multidiffusion} to synthesize ultra-high resolution textures, achieving remarkable results; see an example of size 4096$\times$4096 in Fig.~\ref{fig:4096}. The sampling is conducted over 50 steps using DDIM sampling without classifier-free guidance. At each sampling step, We perform 3 iterations of latent optimization utilizing our attention distillation loss.

\begin{figure}[!ht]
    \centering
    \includegraphics[width=\linewidth]{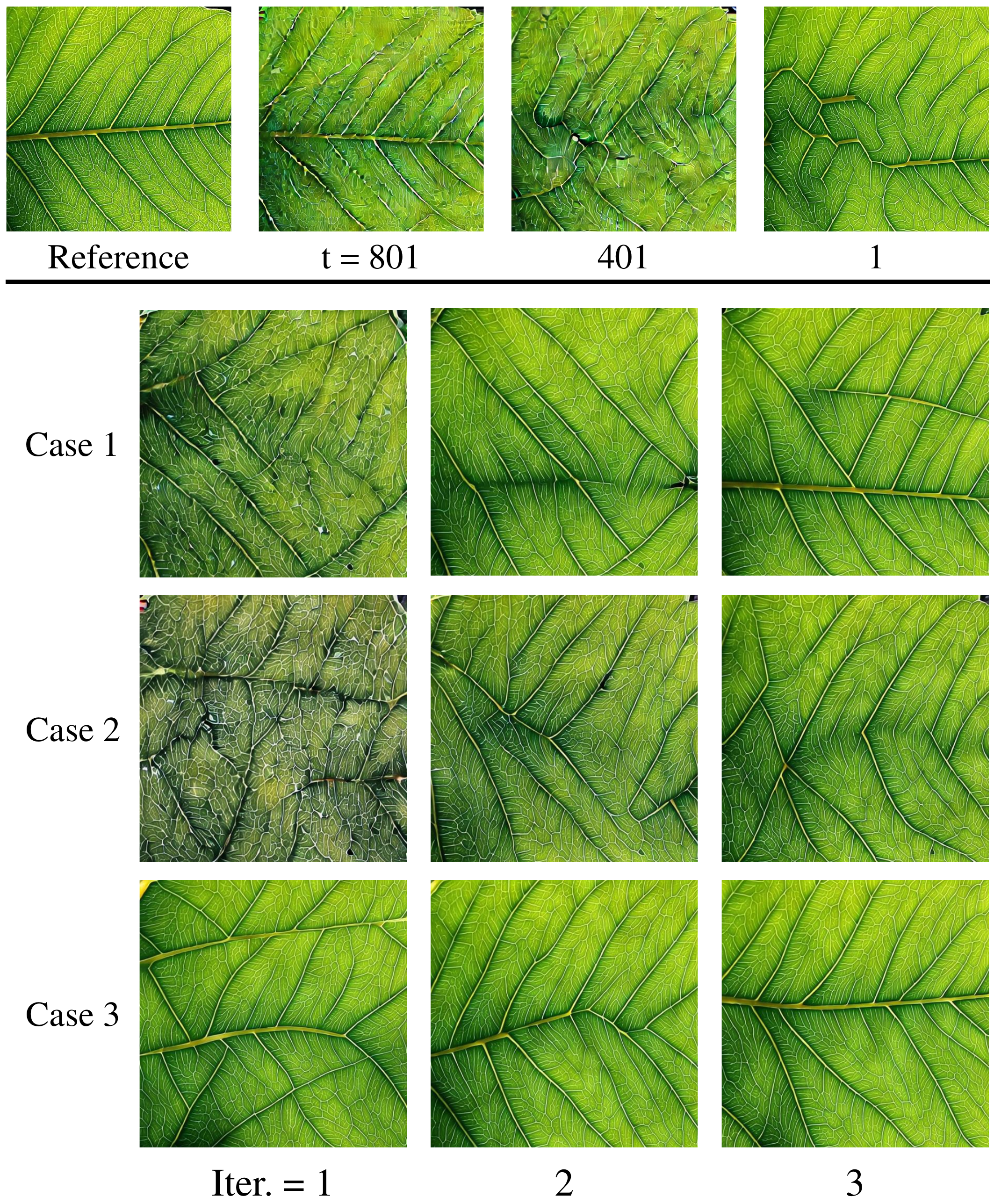} 
    \caption{Digging deeper into the difference between our optimization and sampling-based methods. Top: optimization using differently fixed timesteps (fixed during optimization). Bottom: optimization with clean latents (Case 1), optimization with noised latents (Case 2), and sampling with noised latents (Case 3). For a fair comparison, we add the iteration number at each timestep for the optimization-based method (Case 1 \& 2). See text for details.} 
    \label{fig:distinct2}
\end{figure}

\section{Additional Experiments}
\label{sec:addexp}

\paragraph{Time Efficiency.}
For texture synthesis, either optimization or sampling can be utilized. We record the time consumed by different methods (excluding the time for model loading, compilation, and image encoding/decoding). Specifically, the sampling method employs the DDIM sampler with 50 steps without classifier-free guidance. The Adam optimizer is set with a fixed learning rate of 0.05 for both methods. Typically, non-stationary textures require more iterations to produce a reasonable spatial structure. The detailed results are presented in Table ~\ref{tab:time} and Fig.~\ref{fig:cmp_two}.

\begin{table}[htbp]%
    \small 
    \caption{
    Time efficiency of our optimization-based and sampling-based approaches using Stable Diffusion v1.5. The sample-based approach performed a total of 50 sampling steps.
    }
    \label{tab:time}
    \vspace*{-5mm}
    \begin{center}
    \begin{tabular}{c|ccc|ccc}
      \hline
      & \multicolumn{3}{c|}{Optimization-based} & \multicolumn{3}{c}{Sampling-based} \\  \hline
      Iterations & 100 & 200 & 300 & 1 & 2 & 3 \\ \hline
      Run Time & 10 s & 21 s & 32 s & 7 s & 10 s & 13 s\\ \hline
      GPU Memory & \multicolumn{6}{c}{4 GB} \\
      \hline
    \end{tabular} 
    \end{center}
\end{table}%

\begin{figure}
    \centering
    \includegraphics[width=\linewidth]{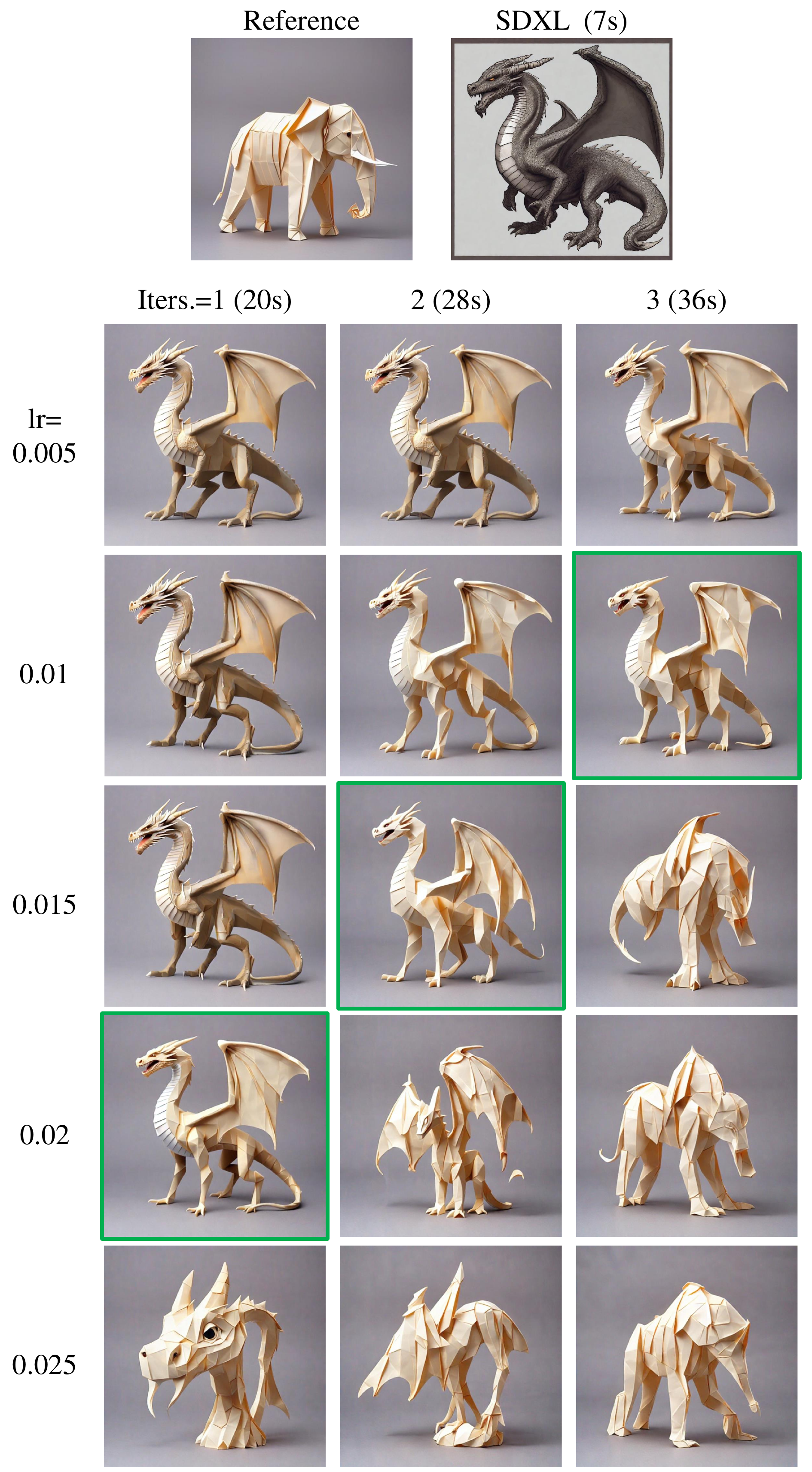} 
    \caption{Impact of different learning rates and optimization iterations in style-specific text-to-image generation.}
    \label{fig:abl_lr_iter}
\end{figure}

\begin{figure}[t]
    \centering
    \includegraphics[width=\linewidth]{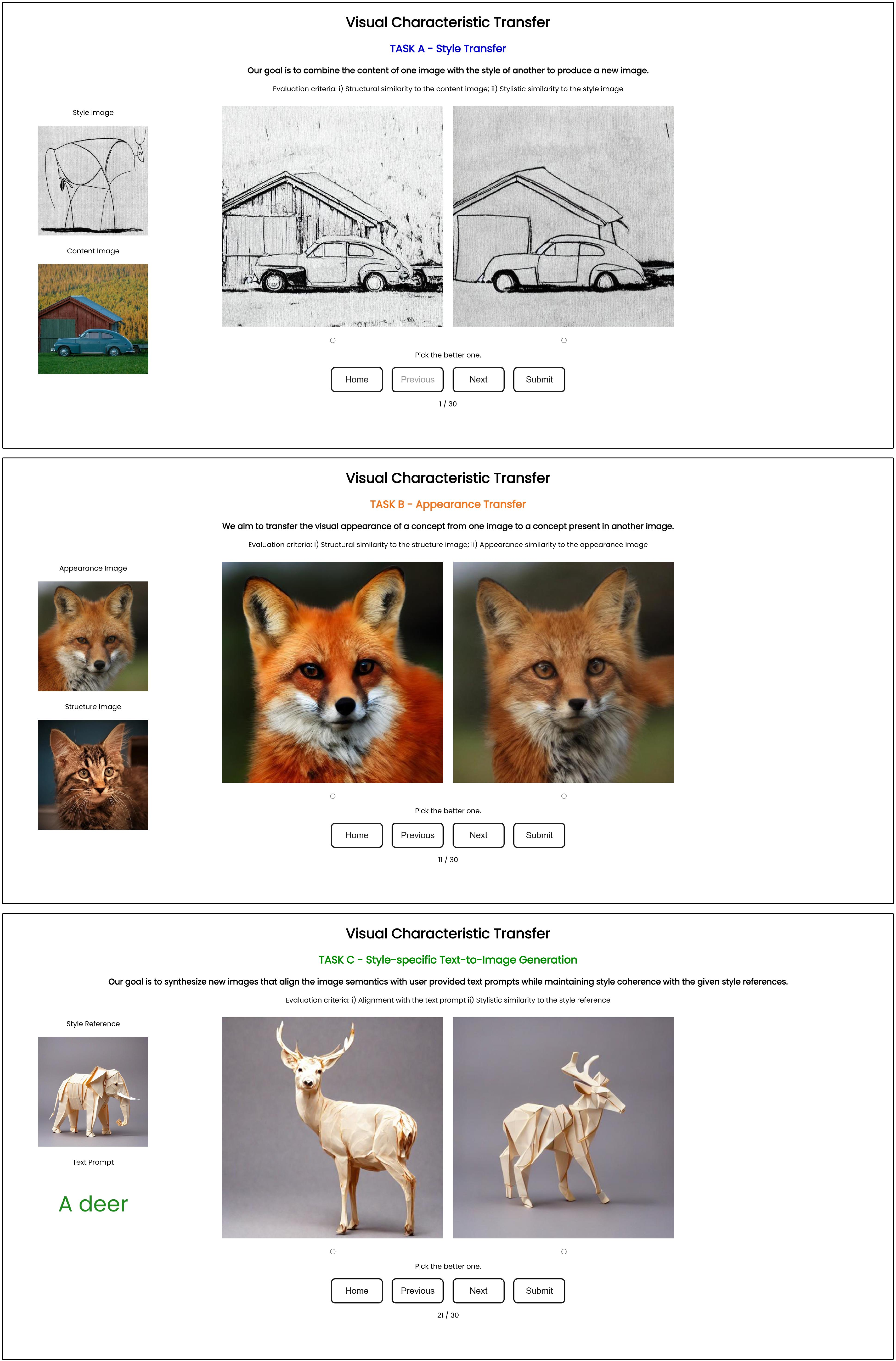} 
    \caption{\textbf{User study interface.}} 
    \label{fig:user_gui}
\end{figure}

\paragraph{A DEEP COMPARISON between optimization and sampling with attention distillation.}
The primary distinction between our optimization-based and sampling-based approaches with attention distillation lies in the nature of the extracted features. As illustrated in Algorithms 1 and 2, sampling-based methods extract features from stochastically inverted images, where scheduled noise relating to timesteps is added, which prevents the latents from being optimized outside the data distribution. In contrast, our optimization-based method extracts features from clean images (\ie, $z_0$ encoded by VAE encoder). Experimental results reveal that by adjusting the timestep $t$, it is possible to extract features at varying levels of granularity, ranging from coarse to fine. As shown in Fig.~\ref{fig:distinct2} top, features extracted from different timesteps were used to compute the AD loss to optimize the same Gaussian noise. Using the Adam optimizer with a learning rate of 0.05 and 200 iterations, the results indicate that features corresponding to larger timesteps focus on coarse structures, whereas those from small timesteps focus on fine details, demonstrating the necessity of linearly decreasing the timestep in our optimization-based method, as described in Sec.3.2 of our main paper.

To further investigate the differences between these two approaches, we designed three experimental cases for texture synthesis. Case 1 involves using features extracted from clean image latents to compute the AD loss for optimization. Case 2 uses features extracted from noisy latents for the same purpose. Case 3 also employs features from noisy latents but optimizes the latent after denoising with the UNet for each timestep, \ie, our AD-guided sampling method. In these experiments, the same Gaussian noise was used as the initial latent, the Adam optimizer with a learning rate of 0.05 was employed, and the number of steps was set to 50. As shown in the bottom of Figure~\ref{fig:distinct2}, the comparison between Cases 1 and 2 reveals that Case 2 produces noisier results and converges much more slowly. More importantly, the comparison between Cases 2 and 3 demonstrates that our guided sampling (or, equivalently, optimizing the denoising UNet-sampled results with AD loss) significantly improves the quality and speed of texture synthesis.

\paragraph{Impact of hyperparameters on Style-Specific T2I Generation.}
We study the impact of two hyperparameters, optimization iteration number in sampling, and learning rate on style-specific T2I generation. As shown in Fig.~\ref{fig:abl_lr_iter}, a lower learning rate or fewer optimization iterations results in insufficient stylization, while increasing the learning rate or the number of optimization iterations can lead to a loss of semantic structure derived from text prompts. According to this study, we set the number of optimization iterations to 2 and the learning rate to 0.015 as default values, balancing image quality, text alignment, and time.

\begin{figure}[t]
    \centering
    \includegraphics[width=\linewidth]{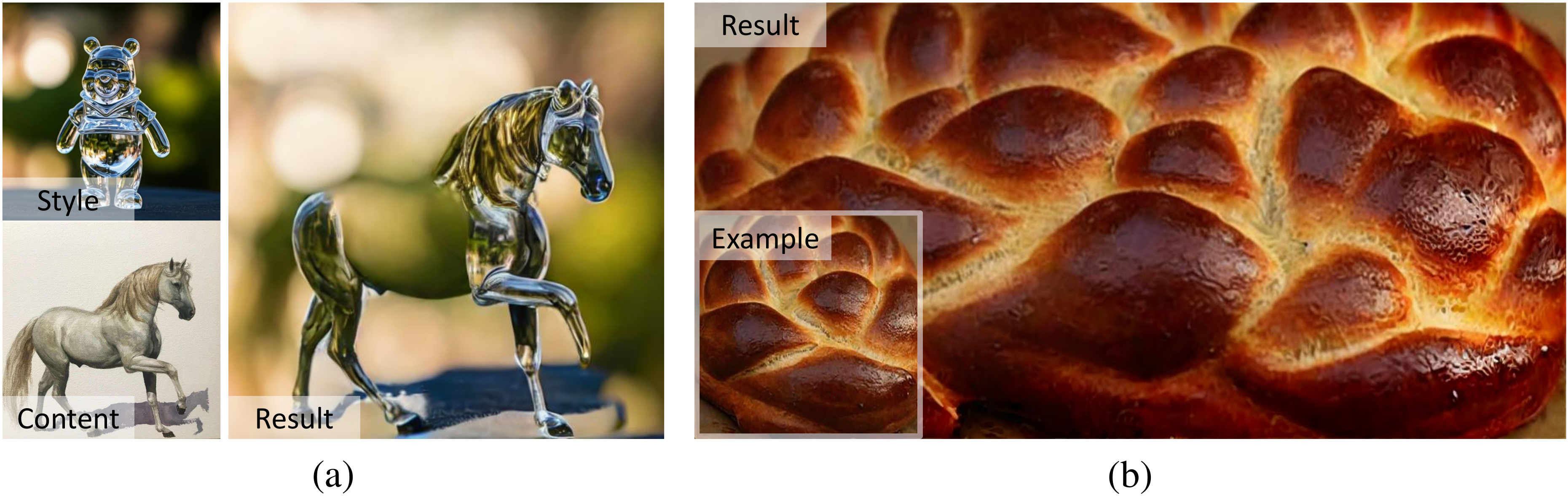} 
    \caption{Limitations of our method.} 
    \label{fig:limitation}
\end{figure}




\section{Details of User Study}
We conduct a user study on three transfer tasks, selecting two competitors for each. Specifically, we compare StyleID~\cite{chung2024styleid} and StyTR$^2$~\cite{Deng_Tang_Dong_Ma_Pan_Wang_Xu_2022} for style transfer, Cross-image Attention~\cite{alaluf2024cross} and SpliceViT~\cite{Tumanyan_Bar-Tal_Bagon_Dekel_2022} for appearance transfer, and InstantStyle~\cite{wang2024instantstyle} and Visual Style Prompting~\cite{jeong2024visual} for style-specific text-to-image generation. For each task, the user interface, shown in Fig.~\ref{fig:user_gui}, randomly presents a set of results from our pool, displaying our method's generated results alongside those of one competitor in the center of the screen side-by-side. Reference images or prompts are provided on the left, with a summary of the evaluation criteria at the top of the screen. Users are asked to pick the better one. The criteria for each task are summarized as follows:

\textbf{Style transfer}: i) structural similarity to the content image, and ii) stylistic similarity to the style image.

\textbf{Appearance transfer}: i) structural similarity to the structure image, and ii) appearance similarity to the appearance image.

\textbf{Style-specific text-to-image generation}: i) semantic alignment with the text prompt, and ii) stylistic similarity to the style reference.


\section{Limitation and Discussion}
While we have demonstrated the effectiveness of our attention distillation loss across a wide range of visual characteristic transfer tasks—such as artistic style and appearance transfer, style-specific text-to-image generation, and texture synthesis—several limitations should be noted. First, we observed that the results of texture expansion occasionally exhibit oversaturated colors. This issue arises because the AD loss does not explicitly constrain the consistency of the data distribution. Instead, it relies on the model's understanding of the reference image to reassemble visual elements. When the resolution of the generated image exceeds the model’s training scope, the aggregation process may produce suboptimal results. Second, in style and appearance transfer tasks, the AD loss depends on the model's ability to establish semantic correspondences based on its understanding of images. When the content of two images differs significantly, the model's limitations may lead to incorrect semantic matches, negatively impacting the final output. See Fig.~\ref{fig:limitation} for two examples.


\section{Additional Results}
\label{sec:moreresults}

\begin{figure*}
    \centering
    \includegraphics[width=\linewidth]{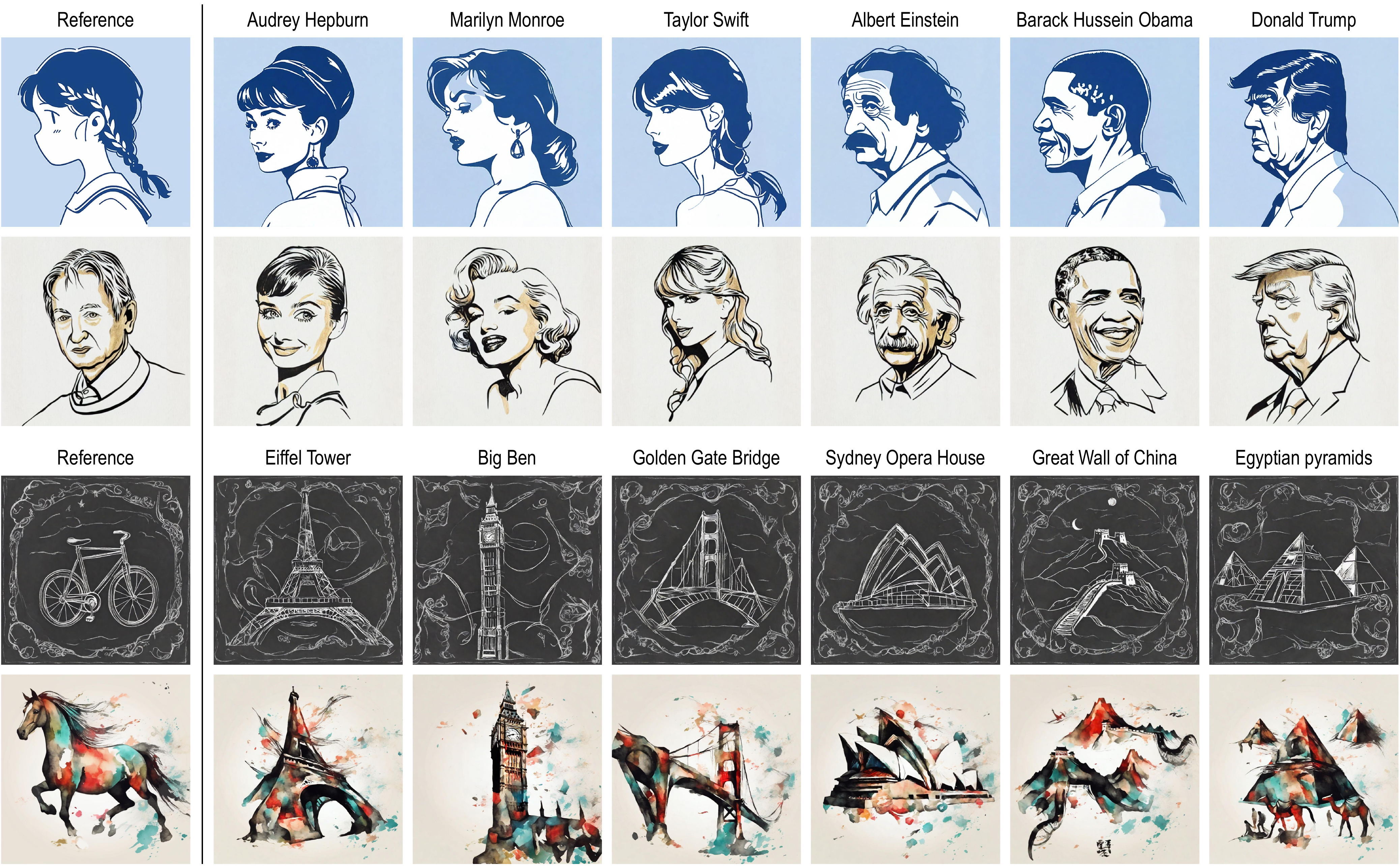} 
    \caption{Additional results of our approach on style-specific text-to-image generation.}
    \label{fig:t2i1}
\end{figure*}

\begin{figure*}
    \centering
    \includegraphics[width=\linewidth]{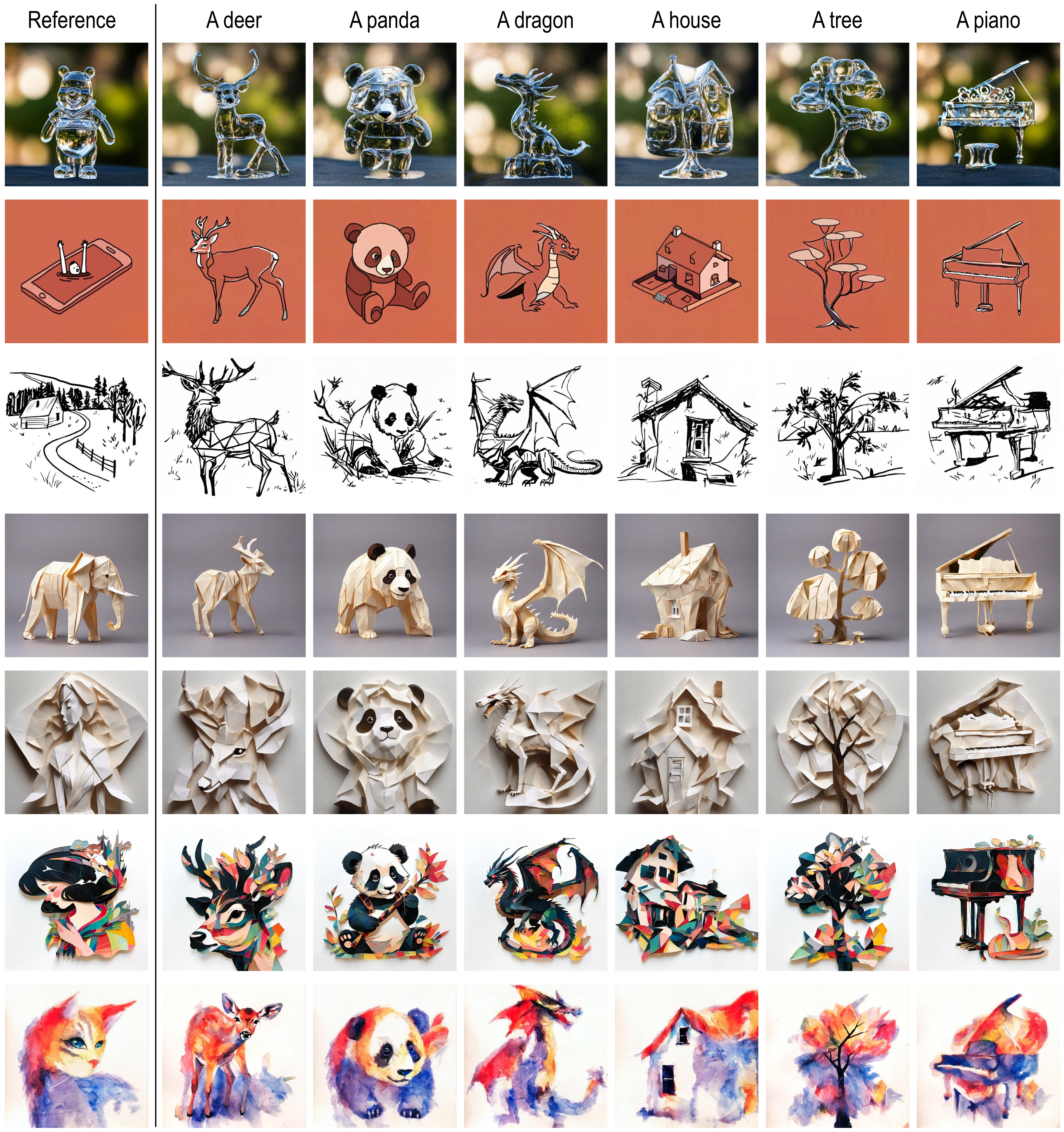} 
    \caption{Additional results of our approach on style-specific text-to-image generation.}
    \label{fig:t2i2}
\end{figure*}

\begin{figure*}
    \centering
    \includegraphics[width=\linewidth]{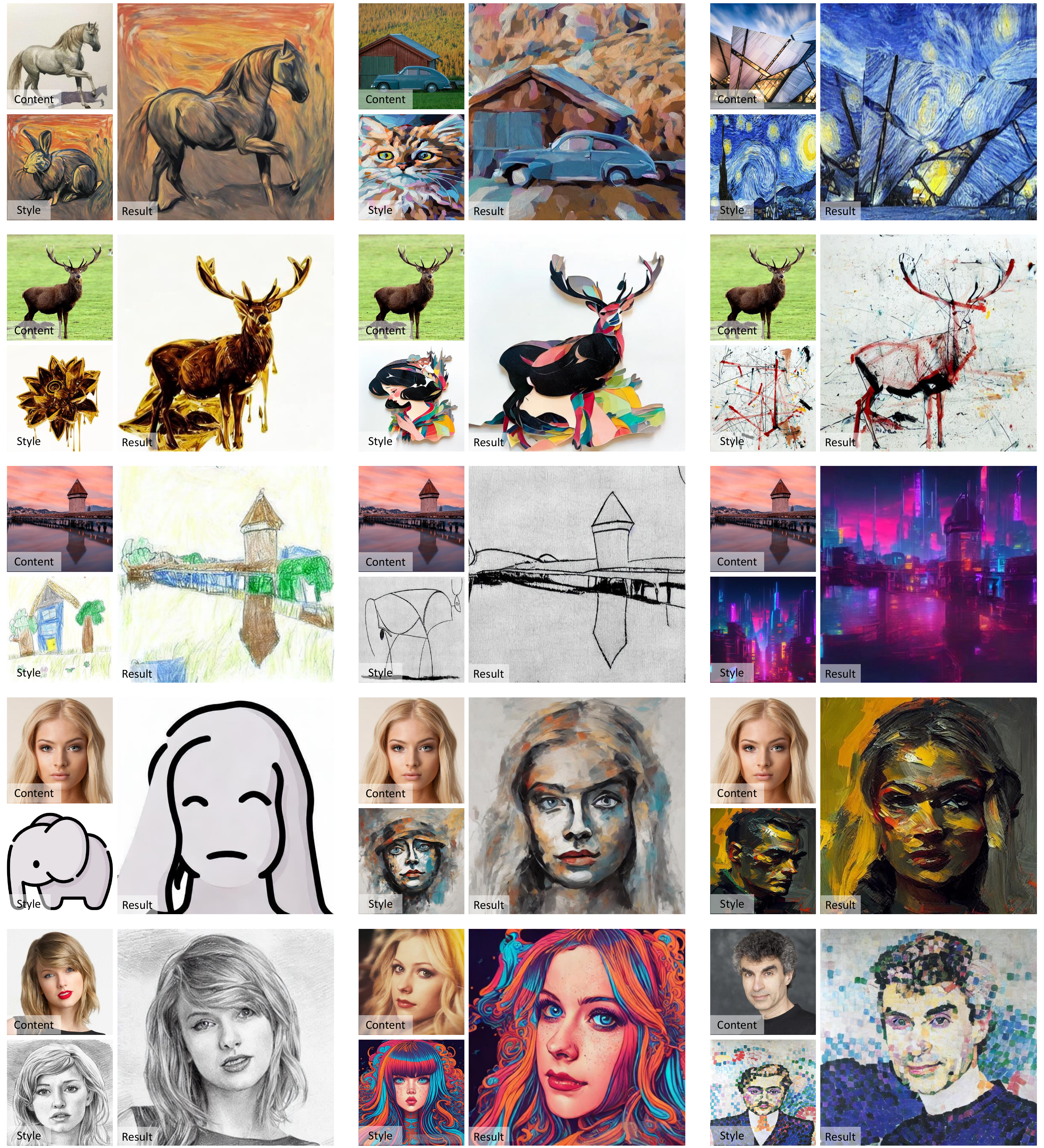} 
    \caption{Additional results of our approach on artistic style transfer.}
    \label{fig:add_transfer}
\end{figure*}

\begin{figure*}
    \centering
    \includegraphics[width=\linewidth]{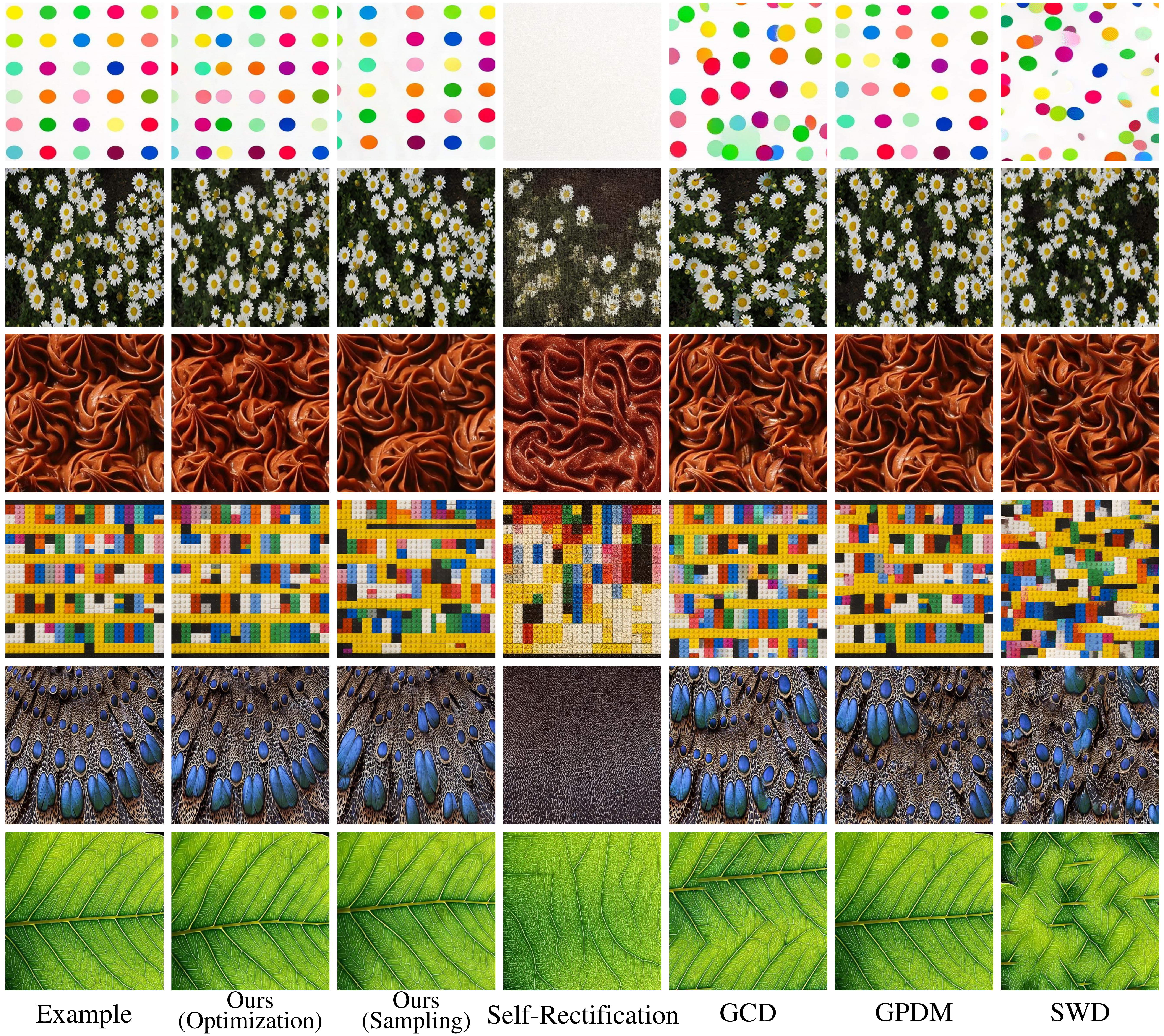} 
    \caption{Comparison on unconditioned texture synthesis. Note that Self-Rectification needs a rough layout, but here, we only give it a random initialization as the target. In our results presented in the 4th and 6th rows, a fine-tuned VAE decoder is employed.}
    \label{fig:add_texcomp}
\end{figure*}

\begin{figure*}
    \centering
    \includegraphics[width=\linewidth]{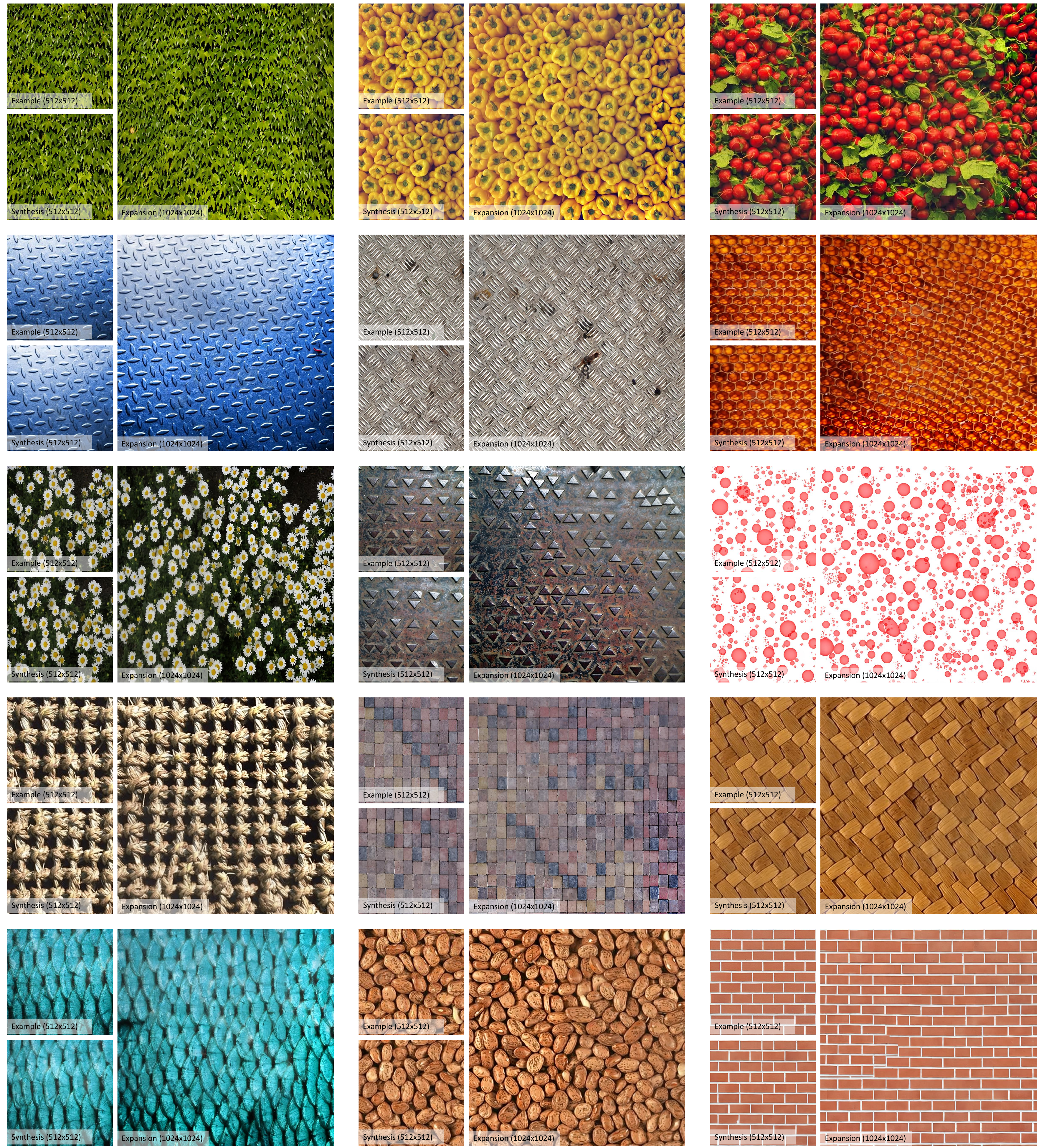} 
    \caption{Additional results of our approach on stationary texture synthesis and expansion.}
    \label{fig:tex1}
\end{figure*}

\begin{figure*}
    \centering
    \includegraphics[width=\linewidth]{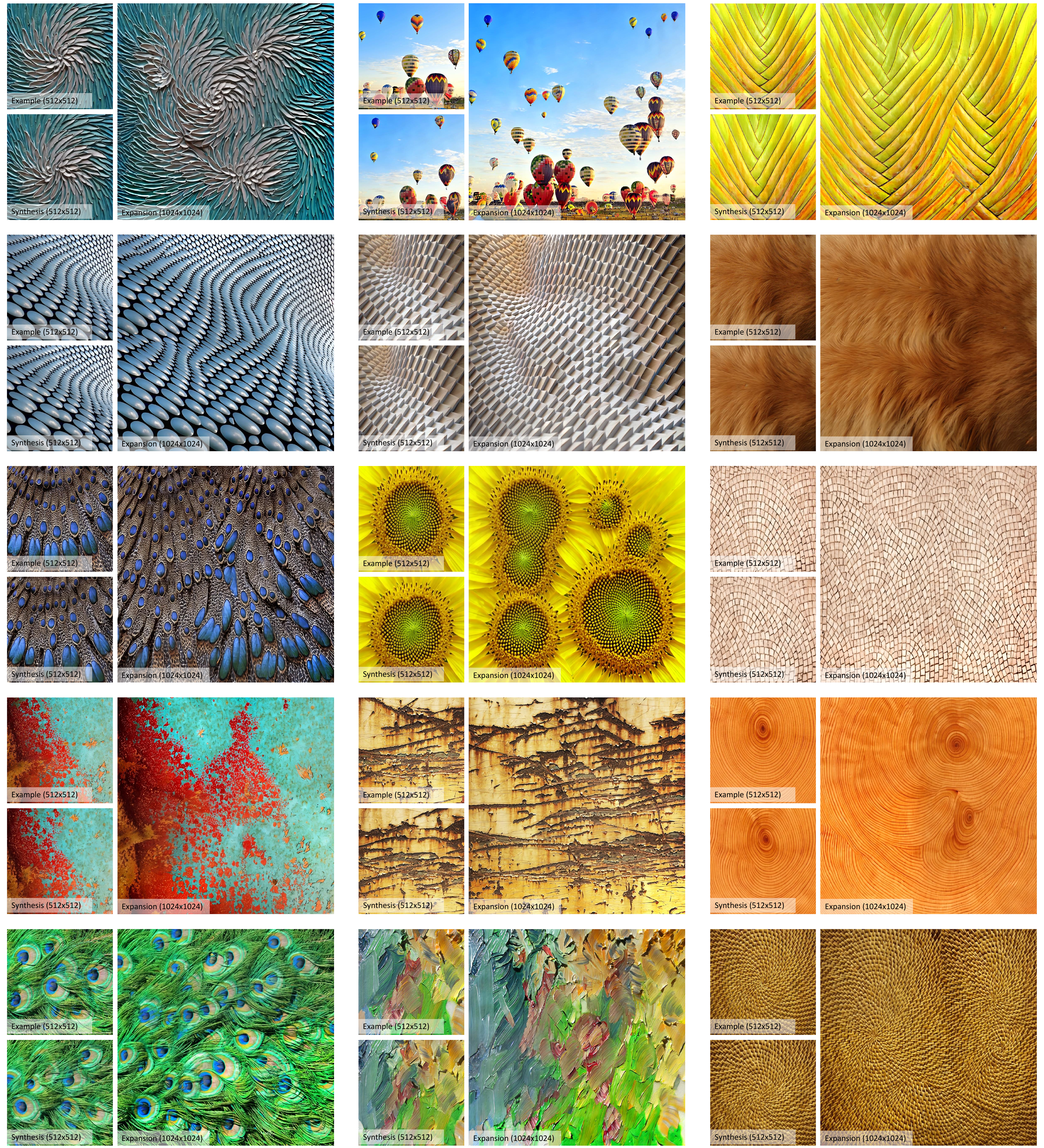} 
    \caption{Additional results of our approach on non-stationary texture synthesis and expansion.}
    \label{fig:tex2}
\end{figure*}

\begin{figure*}
    \centering
    \includegraphics[width=\linewidth]{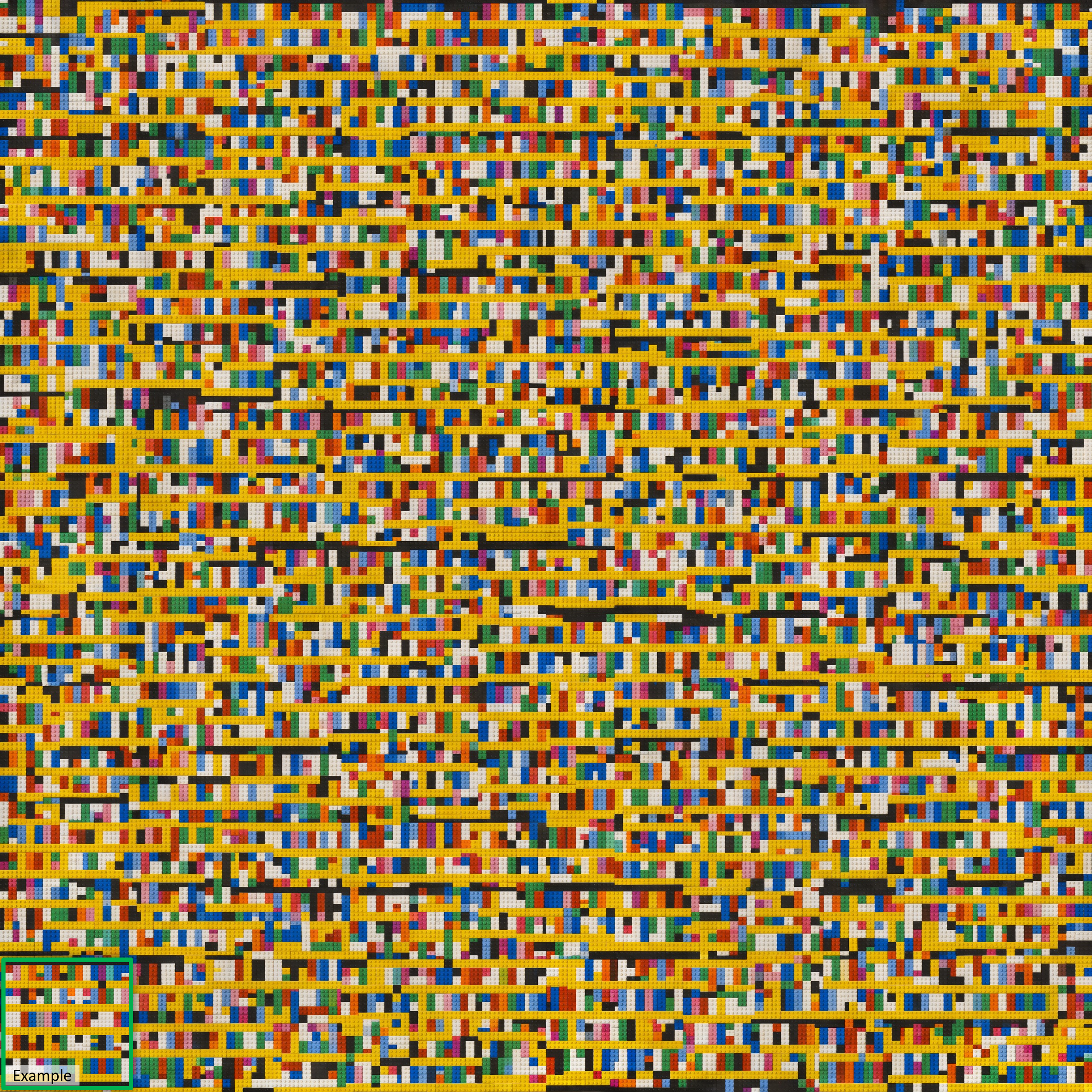} 
    \caption{Texture synthesis with arbitrary resolution, where the above-generated image is in size of 4096$\times$4096 pixels, synthesized from an example (bottom left) in size 512$\times$512.}
    \label{fig:4096}
\end{figure*}

Finally, in the below figures, we provide additional results:

\begin{enumerate}[(1)]
    \item In Figs.~\ref{fig:t2i1} and \ref{fig:t2i2}, we display additional results of creative, text-guided generation with style-specific guidance. 
    \item In Fig.~\ref{fig:add_transfer}, we show more style transfer outcomes on diverse content and style examples.
    \item In Fig.~\ref{fig:add_texcomp}, we present the comparison on unconditioned texture synthesis to showcase the texture understanding capabilities of our attention distillation loss. We apply both optimization-based and sampling-based approaches with our method and compare them against state-of-the-art methods, including Self-Rectification~\cite{zhou2024generating}, GCD~\cite{zhou2023neural}, GPDM~\cite{elnekave2022generating}, and SWD~\cite{heitz2021sliced}.
    \item In Figs.~\ref{fig:tex1} and \ref{fig:tex2}, we present the additional results of stationary and non-stationary texture synthesis and expansion, all achieved through our guided-sampling approach.
    \item Finally, in Fig.~\ref{fig:4096}, we demonstrate an extreme texture expansion by generating a high-resolution image in size 4096$\times$4096 using a 512$\times$512 example.
\end{enumerate}





\end{document}